\documentclass[letterpaper]{article} 
\usepackage{aaai2026}  
\usepackage{times}  
\usepackage{helvet}  
\usepackage{courier}  
\usepackage[hyphens]{url}  
\usepackage{graphicx} 
\usepackage{natbib}  
\usepackage{caption} 
\usepackage{algorithm}
\usepackage{algorithmic}
\usepackage{booktabs}
\usepackage{makecell}
\usepackage{pifont} 
\usepackage{makecell}
\usepackage{algorithm}
\usepackage{algorithmic}
\usepackage{graphicx}
\PassOptionsToPackage{table,dvipsnames}{xcolor}
\usepackage{xcolor}
\usepackage{amsmath, amssymb}    
\usepackage{longtable}
\usepackage{placeins}
\usepackage{multirow}
\usepackage{array}
\usepackage{float}
\usepackage{pifont}
\usepackage{graphicx}
\usepackage{threeparttable}
\usepackage{colortbl} 
\usepackage{newfloat}
\usepackage{listings}
\DeclareCaptionStyle{ruled}{labelfont=normalfont,labelsep=colon,strut=off} 
\floatstyle{ruled}
\newfloat{listing}{tb}{lst}{}
\floatname{listing}{Listing}
\title{Mitigating Perception Bias: A Training-Free Approach to Enhance LMM for Image Quality Assessment}
\author{
    Baoliang Chen\textsuperscript{\rm 1}\equalcontrib, 
    Siyi Pan\textsuperscript{\rm 1}\equalcontrib, 
    Dongxu Wu\textsuperscript{\rm 1},
    Liang Xie\textsuperscript{\rm 2}\thanks{Corresponding author.},
    Xiangjie Sui\textsuperscript{\rm 3},
    Lingyu Zhu\textsuperscript{\rm 4},
    Hanwei Zhu\textsuperscript{\rm 5}
}
\affiliations{
    \textsuperscript{\rm 1}School of Computer Science, South China Normal University, China\\
    \textsuperscript{\rm 2}School of Computer Science and Technology, Guangdong University of Technology, China\\
    \textsuperscript{\rm 3}Faculty of Data Science, City University of Macau, China\\
    \textsuperscript{\rm 4}School of Computer Science, City University of Hong Kong, China\\
    \textsuperscript{\rm 5}School of Computer Science and Engineering, Nanyang Technological University, Singapore\\
    sypan@m.scnu.edu.cn, blchen@scnu.edu.cn
}

\newcommand{\ie}{\textit{i.e.}}       
\newcommand{\eg}{\textit{e.g.}}

\usepackage{bibentry}

\begin{document}

\maketitle

\begin{abstract}
Despite the impressive performance of large multimodal models (LMMs)  in high-level visual tasks, their capacity for image quality assessment (IQA) remains limited. One main reason is that LMMs are primarily trained for high-level tasks (\textit{e.g.}, image captioning), emphasizing unified image semantics extraction under varied quality. Such semantic-aware yet quality-insensitive perception bias inevitably leads to a heavy reliance on image semantics when those LMMs are forced for quality rating. In this paper, instead of retraining or tuning an LMM costly, we propose a training-free debiasing framework, in which the image quality prediction is rectified by mitigating the bias caused by image semantics. Specifically, we first explore several semantic-preserving distortions that can significantly degrade image quality while maintaining identifiable semantics. By applying these specific distortions to the query/test images, we ensure that the degraded images are recognized as poor quality while their semantics mainly remain. During quality inference, both a query image and its corresponding degraded version are fed to the LMM along with a prompt indicating that the query image quality should be inferred under the condition that the degraded one is deemed poor quality.  This prior condition effectively aligns the LMM’s quality perception, as all degraded images are consistently rated as poor quality, regardless of their semantic variance. Finally, the quality scores of the query image inferred under different prior conditions (degraded versions) are aggregated using a conditional probability model. Extensive experiments on various IQA datasets show that our debiasing framework could consistently enhance the LMM performance.
\end{abstract}

\begin{links}
    \link{Code}{https://barrypan12138.github.io/Q-Debias/}
\end{links}

\section{Introduction}
\label{sec:intro}
\begin{figure}[h]
    \centering
    \includegraphics[width=0.45\textwidth]{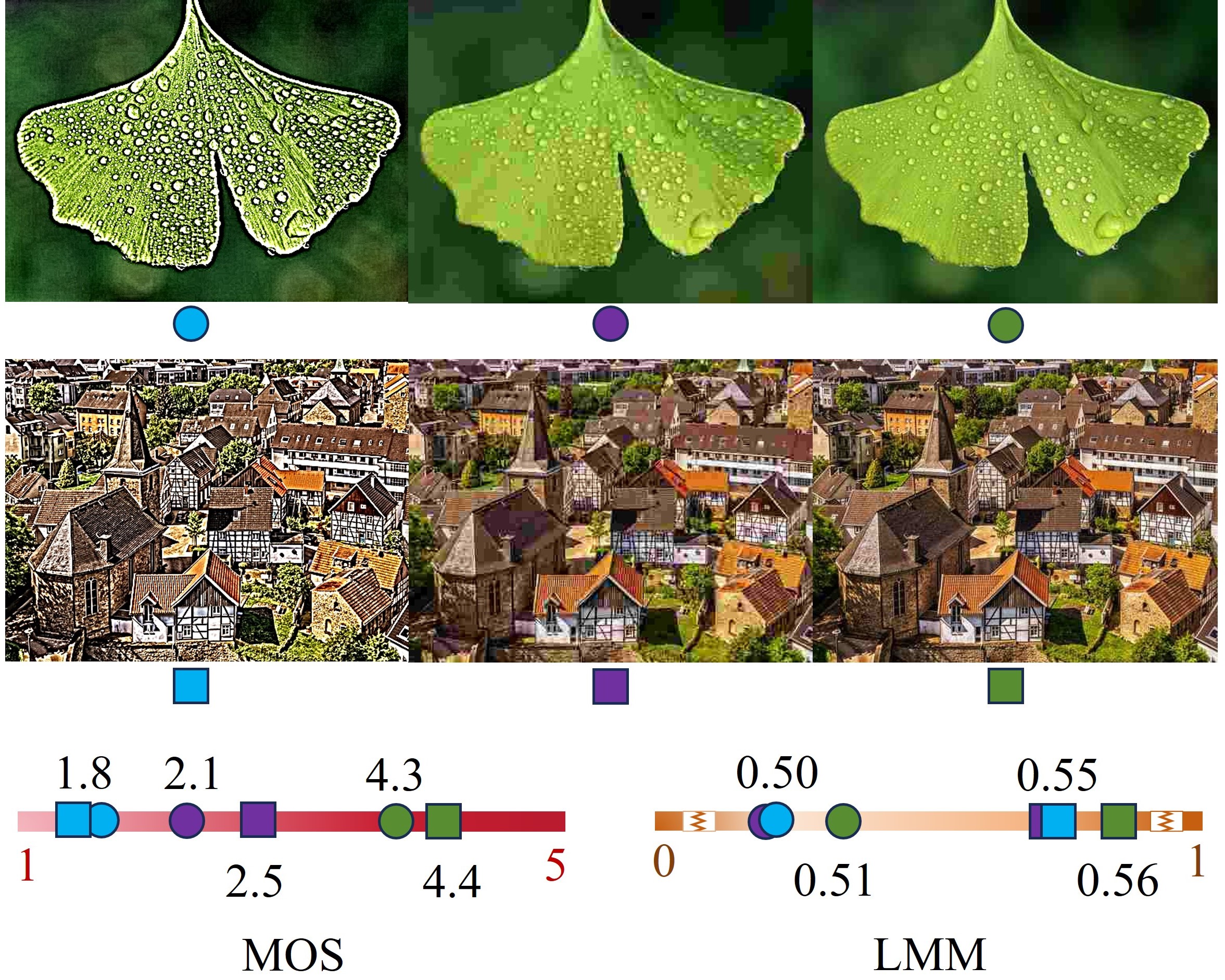}
    \caption{Illustration of perception bias in Large Multimodal Model (LMM) during quality assessment. Image quality ratings from the LMM (mPLUG-Owl3~\cite{ye2024mplugo3}) were obtained using the Q-Bench testing framework~\cite{wu2023qbench}. The LMM consistently assigns higher quality ratings to images in the second row compared to the first, despite both sets exhibiting similar quality distributions as measured by Mean Opinion Scores (MOSs). This discrepancy suggests that the LMM relies more on image semantics than on low-level image clues for quality assessment.}
    \label{fig:intro}
\end{figure}

No-Reference Image Quality Assessment (NR-IQA) models aim to measure image quality in alignment with human perception without any reference, playing a fundamental role across various computer vision tasks \cite{wang2006modern}. 
In the past decades, significant progress has been witnessed in NR-IQA, including traditional hand-crafted feature-based models \cite{chen2025monotonic} and deep learning-based models \cite{chen2021learning}.
Despite the advancement, current models still grapple with limited generalization capabilities on unseen scenes and distortions due to the significant distribution shifts between training and test sets. Recently, the emergence of Large Multimodal Models (LMMs) \cite{openai_gpt4v_2023} has demonstrated impressive generalization abilities across various vision-language tasks, such as classification \cite{wang2023large}, image captioning \cite{yu2023mm}, and visual question answering \cite{shang2024traveler}. However, the focus on high-level visual tasks usually limits their effectiveness in low-level tasks, such as IQA \cite{wu2023qbench, sun2024explore}. Although pioneering researchers have attempted to fine-tune or retrain LMMs for improved IQA accuracy \cite{wu2023align,chen2024q,zhu20242afc}, the laborious dataset construction and costly model training usually render this approach inefficient. Moreover, tuning LMMs specifically for IQA also introduces the risk of catastrophic forgetting \cite{luo2023empirical}, compromising the retention of general knowledge and ultimately degrading the models' capability on other tasks.

A powerful strategy to both retain the LMM’s strengths across tasks and enhance its IQA performance is to unlock its vast general knowledge through well-crafted prompts, encouraging the LMM to respond accurately to quality rating requests.  However, despite this potential, LMMs, driven by training objectives that emphasize semantic extraction over quality, usually default to  \textit{\textbf{interpreting image quality heavily relying on image semantics}}. As shown in Fig.~\ref{fig:intro}, two sets of distorted images with different semantics are presented to an LMM for quality rating. The results indicate that the LMM consistently prefers the quality of images with the second set of semantics, despite both sets having similar quality (Mean Opinion Score (MOS)) distributions. The case reveals that the LMM intrinsically bases its quality rating on image semantics rather than on quality-related clues (\eg, blur). Motivated by this observation, we introduce an innovative, training-free approach to mitigate the perception bias inherent in LMMs. Our enhancement strategy consists of two main steps: 1) \textbf{\textit{bias exposure}}, and  2) \textbf{\textit{bias mitigation}}.

\textbf{\textit{In the bias exposure step}} , we assume the bias exists consistently across images sharing the same semantics. Based on this assumption, we can expose the perception bias of a query/test image by measuring \textbf{\textit{ how much the LMM resists labeling it as high quality when its quality is severely degraded but the semantics are preserved.}}
To achieve this,  we explore several specific distortions that drastically corrupt the image quality while preserving its semantics to some extent. For a query image, we then impose these distortions on the query image and obtain its degraded versions. Herein, those degraded images should be deemed as poor quality. However, the LMM may not always agree with the fact and the disagreement leads to the bias exposure. 

\textbf{\textit{In the bias mitigation step}}, we address the exposed bias using an instructive conditional prompt. Specifically, during inference, we provide both the query image and its degraded version to the LMM, along with a prompt indicating that \textbf{\textit{ the quality of the query image should be assessed under the condition that the degraded counterpart is rated as poor quality}}. By forcing the LMM to rate the quality of the degraded images appropriately, the bias mitigation in turn refines the LMM’s quality prediction for the query image. Finally, the quality predictions under different distortions are aggregated through a conditional probability model, further improving the prediction accuracy. Before delving into detail, we highlight our main contributions as follows:
\begin{itemize} 
\item \textbf{Investigation of Perception Bias in LMM for IQA.} We explore the perception bias inherent in LMM when used for IQA. Our training-free approach, which requires no task-specific fine-tuning, highlights a new pathway for leveraging pre-trained LMM on unseen tasks.

\item \textbf{Conditional Prompt for Bias Mitigation.} We introduce a simple yet effective conditional prompt to mitigate the semantic bias in quality assessment. The prompt encourages the LMM to rate the image quality by aligning the quality prediction of synthetically degraded images, effectively reducing the bias caused by the semantics varies.  Additionally, a confidence-based quality aggregation model is designed, further enhancing the prediction accuracy.

\item \textbf{Comprehensive Evaluation on Diverse Datasets and Distortions.} We extensively evaluated our method on both natural and AI-generated images and the superior performance underscores the high effectiveness of our bias mitigation strategy. In addition, consistent improvements across multiple LMMs also demonstrate the strong generalization of our method, highlighting its potential to successfully extend to future LMMs.

\end{itemize}

\section{Methodology}
\begin{figure*}[h]
    \centering
    \includegraphics[width=\textwidth]{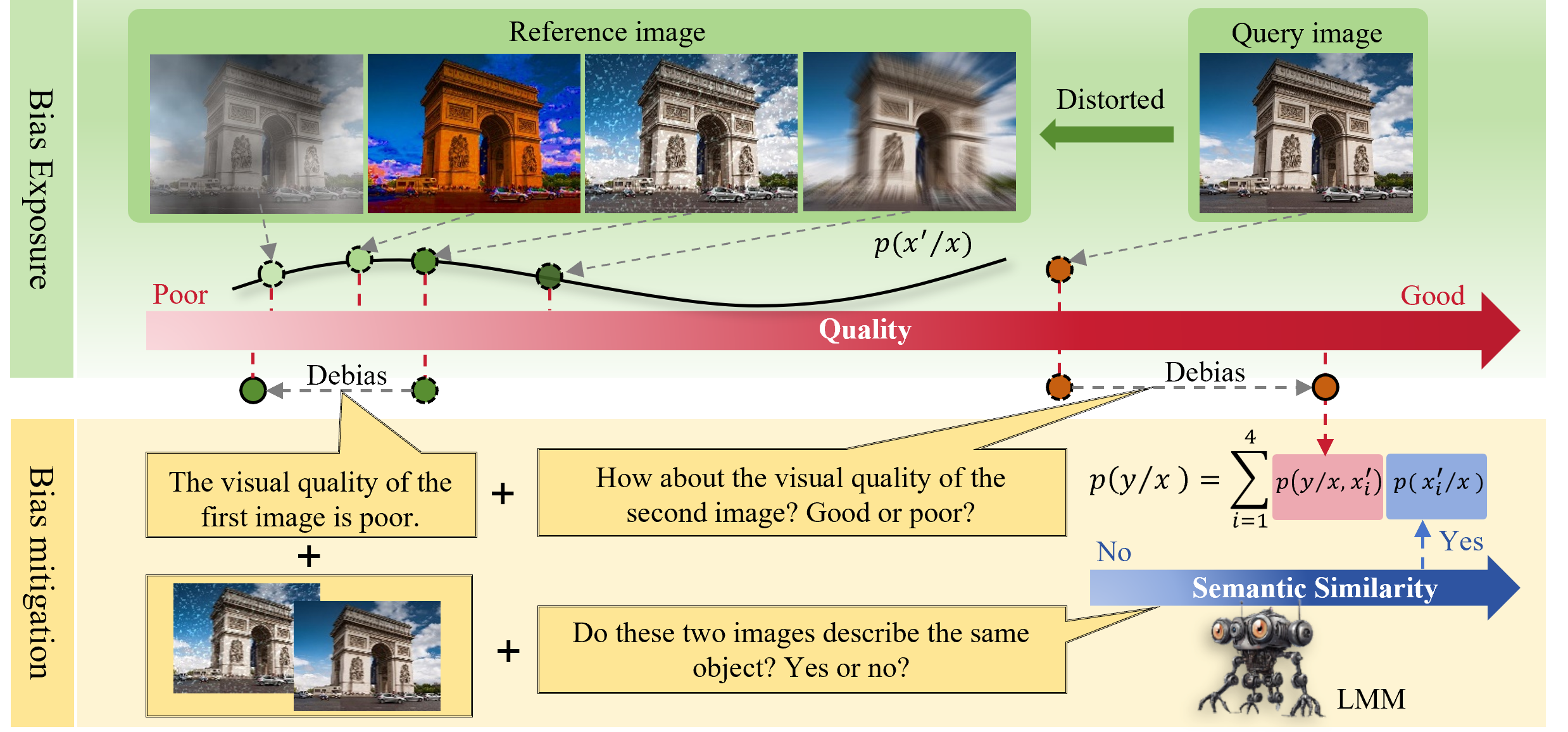}
    \caption{The framework of our perception bias mitigation scheme. It mainly consists of two components: 1) Bias Exposure: Specific distortions are imposed on the query image to significantly degrade the query image quality while preserving its semantics. 
    The disagreement that the LMM rates those distorted images as poor quality exposes the perception bias inherent in the LMM.
    2) Bias Mitigation: Dedicated prompts are defined to mitigate the bias by forcing that the quality of the query image should be assessed under the condition that its degraded counterpart is rated as poor quality. The final quality is then estimated by a semantic similarity based aggregation.}
    \label{fig:architecture}
   
\end{figure*}

\subsection{Preliminary}
Given a query image $x$, the typical prompt for the LMM in IQA is exemplified as follows:
{\fontfamily{cmtt}\selectfont {\noindent \#User: \textcolor{red}{{Rate the quality of the image. Good or poor?} {(Question)} [IMAGE\_TOKEN](Image)} \\\
\noindent \#Assistant: \textcolor{red}{{The quality of the image is [SCORE\_TOKEN].}}}
}

\noindent Based on the predicted logits of `good' token ($x^{\text {gd}}$) and `poor' token ($x^{\text {pr}}$) on the position of  [SCORE\_TOKEN], the image quality score $y$  can be estimated by a  {\fontfamily{cmtt}\selectfont SoftMax} function:
\begin{equation}
p(y \mid x) =\frac{e^{x^{\text {gd }}}}{e^{x^{\text {gd }}}+e^{x^{\text {pr }}}}.
\end{equation}
However, the semantic bias inherent in the LMM usually results in unreliable quality estimation, as the inference heavily relies on image semantics.  To account for this, we adopt a  conditional probability model to mitigate  the bias, which can be formulated as follows,
\begin{equation}
p(y \mid x)=\mathrm{E}_{x^{\prime} \mid x} p\left(y \mid x, x^{\prime}\right) p\left(x^{\prime} \mid x\right),
\label{eqn:pyxx}
\end{equation}
where $x^{\prime}$ is a ``conditional image" 
of $x$, whose quality has been severely degraded while retaining similar semantics to  $x$.  $p\left(x^{\prime} \mid x\right)$ represents the probability distribution of the potential degradation results. During inference, both the conditional image and the query image are fed to the LMM with a prompt instructing the LMM to rate the quality of the query image, under the condition that the conditional image is considered of poor quality.
Our design philosophy is to guide the LMM toward confidently and accurately classifying the degraded images as poor quality, reducing its high reliance on image semantics in quality inference.
This bias mitigation can, in turn, be propagated to the query image quality inference, assuming that the bias is consistently present in images with similar semantics but varying distortions.

\subsection{Framework of Perception Bias Mitigation }
Guided by the model constructed in Eqn.~(\ref{eqn:pyxx}), we design our framework mainly comprises two components: 1) Bias Exposure. Specific distortions that significantly degrade image quality while preserving semantics are explored and imposed on the query image to construct different conditional images $p\left(x^{\prime} \mid x\right)$. 2) Bias Mitigation. A dedicated prompt is designed to estimate $p\left(y \mid x, x^{\prime}\right)$  across different conditional images and obtain the final quality by Eqn.~(\ref{eqn:pyxx}). Our framework is illustrated in Fig.~\ref{fig:architecture} and each component is detailed as follows.\\
\subsubsection{Bias Exposure} 
Given a query image, we examine four typical distortions—zoom blur, spatter noise, saturation enlargement, and fog corruption—to effectively degrade the image quality while preserving its semantic content. Specifically:\\

\noindent \textbf{Zoom Blur.} The zoom blur distortion usually occurs when a camera moves toward an object rapidly, which can be simulated by
\begin{equation}
x^{\prime}_{\text {1}}(u, v)=\frac{1}{n} \sum_{i=1}^n x_{z_i}(u, v),
\label{eqn:zoom}
\end{equation}
where $x_{z_i}$ means the  zoom result of the query image $x$ by a factor $z_i$ and a total of $n$  factors are adopted. $x^{\prime}_{\text {0}}(u, v)$ means the zoom blur results at position $(u,v)$. \\

\noindent \textbf{Spatter Noise.} We use spatter noise to mimic the distortion caused by an unclean camera lens due to bad weather conditions such as rain, mud, or dust, which can be generated by
\begin{equation}
x_2^{\prime}(u, v)=x(u, v) \cdot(1-M(u, v))+C(u, v) \cdot M(u, v),
\end{equation}
where $M(u, v)$ is the spatter mask indicating regions that are affected or unaffected, and $C(u, v)$  represents the specific color distribution adapted for different splatter types. Herein, we use the implementation in \cite{hendrycks2019robustness} to generate \( M(u, v) \) and \( C(u, v) \).\\

\begin{figure*}
    \centering
    \includegraphics[width=\textwidth]{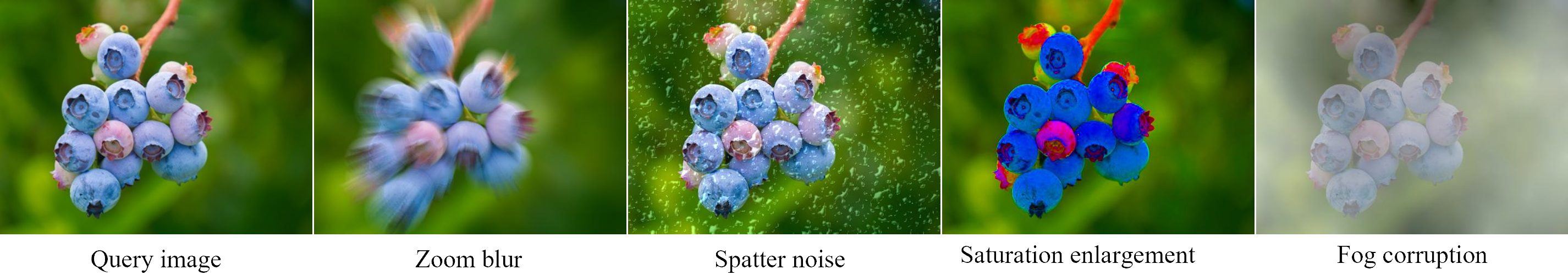}
   \caption{Illustration of the four distortion types which could degrade image quality significantly while largely preserving its semantics.}
    \label{fig:dist}
   
\end{figure*}

\noindent \textbf{Saturation Enlargement.} The saturation distortion modifies the saturation channel of an image in the HSV color space based on the severity parameter $c_0$, which can be defined by
\begin{equation}
x^{\prime}_{\text {3}} = f_{hsv2rgb} (x_h, x^{\prime}_s, x_v),
\end{equation}
with
\begin{equation}
x^{\prime}_s = \operatorname{clip}\left(x_s \cdot c, 0,1\right),
\label{eqn:sat}
\end{equation}
where $x_h, x_s$, and $x_v$ represent the hue, saturation, and value components of $x$ in HSV  space, respectively. $x^{\prime}_s$ denotes the distorted saturation. The function $f_{hsv2rgb}(\cdot)$ converts the image from HSV to RGB space, while  $\operatorname{clip}(\cdot)$ ensures that the distorted results are clipped to a valid range.\\

\noindent \textbf{Fog Corruption.} We simulate a foggy environment by applying a haze effect to the query image as follows: 
\begin{equation}
x^{\prime}_{\text {4}} = \operatorname{clip}\left(x +k \cdot x^{F}, 0, 1 \right), 
\label{eqn:fog}
\end{equation}
where the $ x^{F}$ is the fog pattern generated by a diamond-square algorithm \cite{fournier1982computer,hendrycks2019robustness} based on $x$. $k$ is the hyper-parameter of distortion severity. As shown in Fig.~\ref{fig:dist}, the four types of distortion degrade the image quality significantly while the semantics are still recognizable, leading to an expected bia exposure when they are not deemed poor quality by the LMM.

\subsubsection{Bias Mitigation} 
Based on the generated conditional images for each query image, we then input the query image ($x$) and one of its counterparts ($x^{\prime}_i$) into the LMM,  using a specific prompt to propagate the bias mitigation effect from the conditional image to the query image.

{\fontfamily{cmtt}\selectfont{
\noindent \#User: \textcolor{red}{The visual quality of the first image is poor. How about the visual quality of the second image? Good or poor? {{(Question)}} [IMAGE\_TOKEN1, IMAGE\_TOKEN2].{(Image1, Image2)}} \\
\noindent \#Assistant: \textcolor{red}{ The quality of the image is [SCORE\_TOKEN].}}
}

\noindent Then, the conditional quality probability can be estimated as follows:
\begin{equation}
p(y \mid x, x^{\prime}_i) =\frac{e^{x^{\text {gd }}}}{e^{x^{\text {gd }}}+e^{x^{\text {pr }}}}.
\end{equation}
 Finally, we aggregate the quality estimation across the four distortion types: 
\begin{equation}
p(y \mid x)=\sum_{i=1}^{4}  p\left(y \mid x, x^{\prime}_i\right) p\left({x_i}^{\prime} \mid x\right),
\label{eqn:aggr}
\end{equation}
where \( p\left(x^{\prime}_i \mid x\right) \) is the probability that the distorted image is adopted as the condition. We leverage the semantic similarity between \( x \) and \( x^{\prime} \) to estimate this probability, based on the assumption that the more semantic information maintained, the more confidently the image can be considered as a condition. We achieve the semantic similarity estimation by feeding another prompt to the LMM as follows,
{\fontfamily{cmtt}\selectfont{
\noindent \#User: \textcolor{red}{Do these two images describe the same object? Yes or no? {\fontfamily{cmtt}\selectfont {(Question)}}} \\
\noindent \#Assistant: \textcolor{red}{[SCORE\_TOKEN].}}
}
\noindent This yields 
\begin{equation}
p\left({x_i}^{\prime} \mid x\right) = \frac{e^{w_i}}{\sum_{i=1}^{4} {e^{w_i}}},
\end{equation}
with  
\begin{equation}
w_i =\frac{e^{x_i^{\text {yes }}}}{e^{x_i^{\text {yes }}}+e^{x_i^{\text {no}}}}.
\end{equation}

\section{Experiments}
\begin{figure*}
    \centering
    \includegraphics[width=\textwidth]{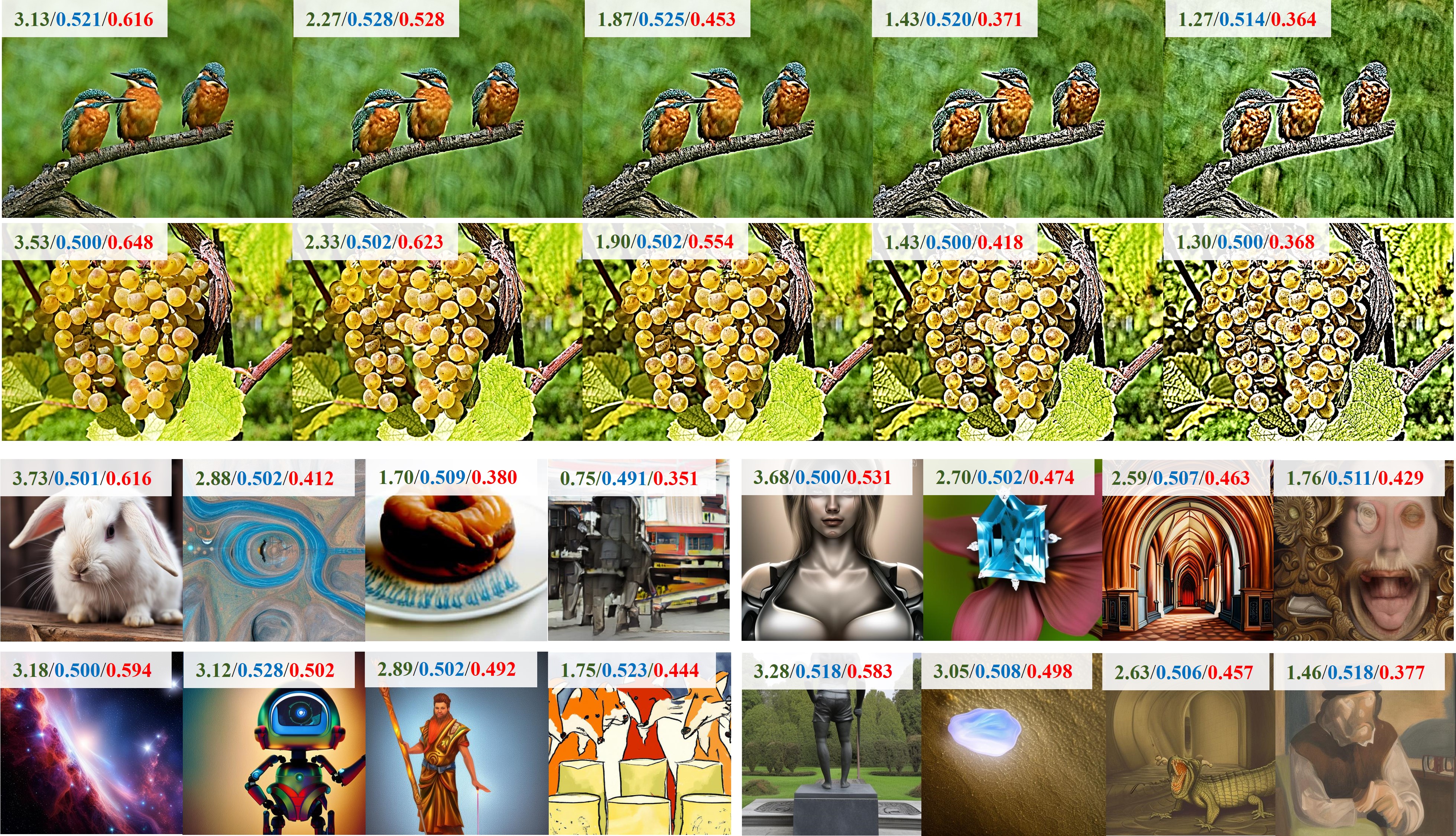}
   \caption{Visualization of image quality prediction results. In each subfigure, the top-left label shows numbers in green, blue and red, representing the {MOS}, the {LMM prediction result with the prompt in Q-Bench} and {our result}, respectively.}
    \label{fig:lmmimg}
\end{figure*}

\subsection{Experimental Settings}
\textbf{Datasets.} We evaluate our method on five publicly available datasets:  LIVE Challenge~\cite{ghadiyaram2015massive}, KonIQ-10k~\cite{koniq10k}, AGIQA-3k~\cite{10262331}, KADID-10k~\cite{kadid10k},  and SPAQ~\cite{fang2020cvpr}. The KonIQ-10k, SPAQ, and LIVE Challenge datasets are in-the-wild image collections, featuring authentic distortions. KonIQ-10k and SPAQ datasets each contain over 10,000 images and the SPAQ dataset is specifically designed to assess the quality of images captured by smartphones. The KADID-10k dataset comprises 10,125 images with various systematic distortions. The AGIQA-3k  dataset includes 2,900 images focused on AI-generated image quality assessment. \\

\noindent\textbf{Comparison Methods.} We denote our method as ``Q-Debias" and compare its performance against both training-free (opinion-unaware) and training-based methods across multiple datasets. The training-free methods include BLINDS-II~\cite{moorthy2010two}, BRISQUE~\cite{mittal2012no}, NIQE~\cite{mittal2012making}, NPQI~\cite{liu2020blind}, ContentSep~\cite{babu2023no}, CLIP-IQA~\cite{wang2023exploring}, MDFS~\cite{ni2024opinion}, and Q-Bench~\cite{wu2023qbench}. In particular, Q-Bench refers to using the same LMM (mPLUG-Owl3) with our method, while applying the query prompt from Q-Bench. For the training-based methods, we adopt models trained on the large-scale KonIQ-10k dataset for comparison, including ARNIQA~\cite{agnolucci2024arniqa}, TReS~\cite{golestaneh2021no}, and MUSIQ~\cite{ke2021musiq}. We list their performance in cross-dataset settings to verify their generalization capability. The Pearson Linear Correlation Coefficient (PLCC) and the Spearman Rank-Order Correlation Coefficient (SRCC) are used as metrics to assess the linearity and monotonicity of our quality predictions.\\

\noindent\textbf{Implementation Details.} 
We select mPLUG-Owl3 as our LMM due to its superior performance on image processing tasks. We set \( z_i \in \{1.00, 1.01, 1.02, \dots, 1.10\} \), \( n = 11 \) in Eqn.~(\ref{eqn:zoom}), and \( c = 2.0 \) in Eqn.~(\ref{eqn:sat}). We set \( k = 2.5 \) in Eqn.~(\ref{eqn:fog}). 

{
\begin{table*}[t]
\centering
\caption{Performance comparison of our method, Q-Debias, against both training-free and training-based IQA models. The {percentage} indicates the improvement of our method over Q-Bench. The best two results are highlighted in boldface.}
\label{tab1}
\resizebox{\textwidth}{!}{
\begin{threeparttable}
\begin{tabular}{l|c|cc|cc|cc|cc|cc}
\hline
\multirow{2}{*}{{Methods}} & {Training-free?} & \multicolumn{2}{c|}{{LIVE Challenge}} & \multicolumn{2}{c|}{{KonIQ-10k}} & \multicolumn{2}{c|}{{AGIQA-3k}} & \multicolumn{2}{c|}{{KADID-10k}} & \multicolumn{2}{c}{{SPAQ}}\\ 
 \cline{3-12}
  & {(Training set)} & {SRCC}↑ & {PLCC}↑ & {SRCC}↑ & {PLCC}↑ & {SRCC}↑ & {PLCC}↑ & {SRCC}↑ & {PLCC}↑ & {SRCC}↑ & {PLCC}↑  \\ \hline

BLIINDS-II~\cite{moorthy2010two} & \checkmark & 0.090 & 0.107 & 0.585 & 0.598 & 0.454 & 0.510 & 0.224 & 0.313 & 0.317 & 0.326   \\ 
QAC~\cite{xue2013learning} & \checkmark & 0.226 & 0.284 & 0.340 & 0.291 & - & - & 0.239 & 0.309 & 0.440 & 0.450 \\ 
BRISQUE~\cite{mittal2012no} & \checkmark & 0.561 & 0.598 & 0.705 & 0.707 & 0.493 & 0.533 & 0.330 & 0.370 & 0.484 & 0.481  \\ 
NIQE~\cite{mittal2012making} & \checkmark & 0.463 & 0.491 & 0.551 & 0.488 & 0.528 & 0.520 & 0.379 & 0.389 & 0.703 & 0.671 \\ 
ILNIQE~\cite{zhang2015feature} & \checkmark & 0.439 & 0.503 & 0.505 & 0.496 & 0.594 & 0.623 & 0.540 & 0.534 & 0.696 & 0.637   \\ 
NPQI~\cite{liu2020blind} & \checkmark & 0.475 & 0.490 & 0.613 & 0.614 & 0.658 & 0.714 & 0.391 & 0.340 & 0.600 & 0.616  \\ 
ContentSep~\cite{babu2023no} & \checkmark & 0.506 & 0.513 & 0.640 & 0.627 & - & - & 0.506 & 0.357 & 0.708 & 0.665  \\ 
CLIP-IQA~\cite{wang2023exploring} & \checkmark & 0.612 & 0.594 & 0.695 & \textbf{0.727} & 0.658 & 0.714 & 0.500 & 0.520 & 0.738 & 0.735  \\
MDFS~\cite{ni2024opinion} & \checkmark & 0.482 & 0.536 & \textbf{0.733} & 0.712 & \textbf{0.672} & 0.676 & 0.598 & 0.594 & 0.741 & 0.718 \\ \hline
ARNIQA~\cite{agnolucci2024arniqa} & \ding{55} (KonIQ-10k) & 0.670 & 0.715 & - & - & 0.621 & 0.694 & \textbf{0.725} & \textbf{0.717} & 0.576 & 0.577  \\ 
TReS~\cite{golestaneh2021no} & \ding{55} (KonIQ-10k) & 0.771 & 0.805 & - & - & 0.652 & \textbf{0.737} & 0.468 & 0.492 & 0.418 & 0.417  \\
MUSIQ~\cite{ke2021musiq} & \ding{55} (KonIQ-10k) & \textbf{0.788} &\textbf{0.824} & - & - & 0.630 & {0.722} & 0.556 & 0.575 & 0.726 & \textbf{0.738}  \\ \hline
Q-Bench \tnote{*} ~\cite{wu2023qbench} & \checkmark & 0.721 & 0.677 & 0.672 & 0.573 & 0.596 & 0.469 & 0.315 & 0.267 & \textbf{0.767} & 0.650  \\ \hline
\rowcolor{gray!10}
\textbf{Q-Debias} (Ours) & \checkmark & \textbf{0.794} & \textbf{0.818} & \textbf{0.838} & \textbf{0.863} & \textbf{0.717} & \textbf{0.753} & \textbf{0.700} & \textbf{0.753} & \textbf{0.867}  & \textbf{0.826} 
 \\ \hline
\rowcolor{gray!10}
 \multicolumn{2}{l|}{{Improvement over Q-bench}}  & \textbf{$\uparrow$ 10.1\%} & \textbf{$\uparrow$ 20.8\%} & \textbf{$\uparrow$ 24.7\%} &  \textbf{$\uparrow$ 50.6\%} &  \textbf{$\uparrow$ 20.3\%} &  \textbf{$\uparrow$ 60.6\%} &  \textbf{$\uparrow$ 122.2\%} &  \textbf{$\uparrow$ 167.0\%} &  \textbf{$\uparrow$ 13.0\%} &  \textbf{$\uparrow$ 27.1\%}  \\ \hline
\end{tabular}
    \begin{tablenotes}
      \footnotesize
      \item[*] We use the same quality query prompt from Q-Bench and set the LMM as mPLUG-0wl3, consistent with our method.
    \end{tablenotes}
\end{threeparttable}}
\end{table*}


\begin{table*}[htbp]
  \centering
  \caption{Performance improvement on other LMMs.}
  \resizebox{\textwidth}{!}{
    \begin{tabular}{l|cc|cc|cc|cc|cc|cc}
    \toprule
    \multicolumn{1}{c|}{\multirow{3}[6]{*}{Models}} & \multicolumn{4}{c|}{LIVE Challenge} & \multicolumn{4}{c|}{AGIQA-3k} & \multicolumn{4}{c}{Average} \\
\cmidrule{2-13}          & \multicolumn{2}{c|}{Vanilla prompt} & \multicolumn{2}{c|}{Our debias prompt} & \multicolumn{2}{c|}{Vanilla prompt} & \multicolumn{2}{c|}{Our debias prompt} & \multicolumn{2}{c|}{Vanilla prompt} & \multicolumn{2}{c}{Our debias prompt} \\
\cmidrule{2-13}          & \multicolumn{1}{c}{SRCC↑ } & \multicolumn{1}{c|}{PLCC↑ } & \multicolumn{1}{c}{SRCC↑ } & \multicolumn{1}{c|}{PLCC↑ } & \multicolumn{1}{c}{SRCC↑ } & \multicolumn{1}{c|}{PLCC↑ } & \multicolumn{1}{c}{SRCC↑ } & \multicolumn{1}{c|}{PLCC↑ } & \multicolumn{1}{c}{SRCC↑} & \multicolumn{1}{c|}{PLCC↑} & \multicolumn{1}{c}{SRCC↑} & \multicolumn{1}{c}{PLCC↑} \\
    \midrule
    BakLLaVA~\cite{liu2024improved} & 0.090 & 0.108 & 0.263 ( \textbf{$\uparrow$ 192.0\%}) & 0.265 ( \textbf{$\uparrow$ 145.0\%}) & 0.480 & 0.321 & 0.460 & 0.482 ( \textbf{$\uparrow$ 50.2\%}) & 0.285 & 0.215 & 0.362 ( \textbf{$\uparrow$ 27.0\%}) & 0.374 ( \textbf{$\uparrow$ 74.0\%})\\
    Qwen-VL~\cite{bai2023qwen}  & 0.470 & 0.546 & 0.504 ( \textbf{$\uparrow$ 7.20\%}) & 0.501 & 0.504  & 0.532  & 0.615 ( \textbf{$\uparrow$ 22.0\%}) & 0.623 ( \textbf{$\uparrow$ 17.1\%}) & 0.487 & 0.539 & 0.560 ( \textbf{$\uparrow$ 15.0\%}) & 0.562 ( \textbf{$\uparrow$ 4.27\%}) \\
    LLaVA-OneVision~\cite{li2024llavaonevisioneasyvisualtask}  & 0.379 & 0.654 & 0.631 ( \textbf{$\uparrow$ 66.5\%}) & 0.649 & 0.581  & 0.781  & 0.707 ( \textbf{$\uparrow$ 8.94\%}) & 0.806 ( \textbf{$\uparrow$ 14.0\%}) & 0.480 & 0.718 & 0.669 ( \textbf{$\uparrow$ 39.4\%}) & 0.728( \textbf{$\uparrow$ 1.39\%}) \\
    LLaVA-Interleave~\cite{li2024llavanextinterleavetacklingmultiimagevideo}  & 0.221 & 0.337 & 0.454 ( \textbf{$\uparrow$ 100.5\%}) & 0.543 ( \textbf{$\uparrow$ 61.1\%}) & 0.223  & 0.315  & 0.464 ( \textbf{$\uparrow$ 108.1\%}) & 0.560 ( \textbf{$\uparrow$ 77.8\%}) & 0.222 & 0.326 & 0.459 ( \textbf{$\uparrow$ 106.8\%}) & 0.552( \textbf{$\uparrow$ 69.3\%}) \\
    DeepSeek-VL2~\cite{wu2024deepseekvl2mixtureofexpertsvisionlanguagemodels}  & 0.800 & 0.851 & 0.822 ( \textbf{$\uparrow$ 2.75\%}) & 0.860 ( \textbf{$\uparrow$ 1.06\%}) & 0.606  & 0.655  & 0.759 ( \textbf{$\uparrow$ 25.2\%}) & 0.778 ( \textbf{$\uparrow$ 17.0\%}) & 0.703 & 0.753 & 0.791 ( \textbf{$\uparrow$ 12.5\%}) & 0.819( \textbf{$\uparrow$ 8.76\%}) \\
    \bottomrule
    \end{tabular}}%
  \label{tab:lmm}%
\end{table*}%
}
\begin{table}[htbp]
  \centering
  \caption{Ablation study of different types of conditional images. The best results in each setting are highlighted in boldface.}
  \resizebox{0.5\textwidth}{!}{
    \begin{tabular}{c|cc|cc|c|cc|cc|cc}
    \toprule
    \multirow{2}[4]{*}{Exp. ID} & \multicolumn{4}{c|}{\multirow{2}[4]{*}{Single Distortion}} & \multirow{2}[4]{*}{\makecell[c]{Semantic \\ Consistency}} & \multicolumn{2}{c|}{LIVE Challenge} & \multicolumn{2}{c|}{AGIQA-3k} & \multicolumn{2}{c}{Average} \\
\cmidrule{7-12}          & \multicolumn{4}{c|}{}         &       & SRCC↑ & PLCC↑ & SRCC↑ & PLCC↑ & SRCC↑ & PLCC↑\\
    \midrule
     1     & \multicolumn{2}{c|}{\multirow{3}[2]{*}{Blur}} & \multicolumn{2}{c|}{Zoom} & $\checkmark$    & 0.644  & 0.673  & 0.633  & 0.658  & \textbf{0.639} & \textbf{0.666} \\
     2     & \multicolumn{2}{c|}{} & \multicolumn{2}{c|}{Motion} & $\checkmark$    & 0.497  & 0.515  & 0.552  & 0.546  & 0.525  & 0.531  \\
     3     & \multicolumn{2}{c|}{} & \multicolumn{2}{c|}{Gaussian} & $\checkmark$    & 0.617  & 0.497  & 0.636  & 0.648  & 0.627  & 0.573  \\
    \midrule
    4     & \multicolumn{2}{c|}{\multirow{2}[2]{*}{Noise}} & \multicolumn{2}{c|}{Gaussian} & $\checkmark$    & 0.677  & 0.646  & 0.686  & 0.720  & 0.682  & 0.683  \\
    5     & \multicolumn{2}{c|}{} & \multicolumn{2}{c|}{Spatter} & $\checkmark$    & 0.762  & 0.799  & 0.713  & 0.768  & \textbf{0.738} & \textbf{0.784} \\
    \midrule
     6     & \multicolumn{2}{c|}{\multirow{3}[2]{*}{Bad weather}} & \multicolumn{2}{c|}{Snow} & $\checkmark$    & 0.713  & 0.761  & 0.686  & 0.640  & 0.700  & 0.701  \\
     7     & \multicolumn{2}{c|}{} & \multicolumn{2}{c|}{Frost} & $\checkmark$    & 0.632  & 0.705  & 0.633  & 0.573  & 0.633 & 0.639 \\
     8     & \multicolumn{2}{c|}{} & \multicolumn{2}{c|}{Fog} & $\checkmark$    & 0.729  & 0.763  & 0.689  & 0.702  & \textbf{0.709} & \textbf{0.733} \\
    \midrule
    9     & \multicolumn{4}{c|}{Brightness} & $\checkmark$    & 0.613  & 0.673  & 0.620  & 0.668  & \textbf{0.617} & \textbf{0.671} \\
    10    & \multicolumn{4}{c|}{Saturation} & $\checkmark$    & 0.784  & 0.790  & 0.720  & 0.735  & \textbf{0.752} & \textbf{0.763} \\
    \midrule
    \multicolumn{12}{c}{Multiple Distortions} \\
    \midrule
    Exp. ID & Zoom  & \multicolumn{1}{c}{Spatter} & Fog   & Saturation & Semantic & SRCC↑  & PLCC↑  & SRCC↑  & PLCC↑  & SRCC↑  & PLCC↑ \\
    \midrule
    11    & \ding{55}    & \multicolumn{1}{c}{\ding{55}} & \ding{55}    & \ding{55}    & \ding{55}    & 0.721  & 0.677  & 0.596  & 0.469  & 0.659  & 0.573  \\
    12    & \ding{55}    & \multicolumn{1}{c}{$\checkmark$} & \ding{55}    & $\checkmark$    & $\checkmark$    & 0.793  & 0.790  & 0.709  & 0.712  & 0.751  & 0.763  \\
    13    & \ding{55}    & \multicolumn{1}{c}{$\checkmark$} & $\checkmark$    & $\checkmark$    & $\checkmark$    & 0.793  & 0.773  & 0.714  & 0.702  & 0.753  & 0.738  \\
    14    & $\checkmark$    & \multicolumn{1}{c}{$\checkmark$} & $\checkmark$    & $\checkmark$    & \ding{55}    & 0.493  & 0.472  & 0.518  & 0.508  & 0.506  & 0.490  \\
    \midrule
    Q-Debias & $\checkmark$    & \multicolumn{1}{c}{$\checkmark$} & $\checkmark$    & $\checkmark$    & $\checkmark$    & \textbf{0.794} & \textbf{0.818} & \textbf{0.717} & \textbf{0.753} & \textbf{0.756} & \textbf{0.786} \\
    \bottomrule
    \end{tabular}}%

  \label{tab:dst}%
\end{table}%

\subsection{Comparison with NR-IQA Models}
\noindent\textbf{Prediction Accuracy.} As shown in Table~\ref{tab1}, compared with existing training-free methods, our model Q-Debias consistently achieves the best performance across the first five IQA datasets. In particular, most hand-crafted feature-based models (\eg, NIQE) experience a significant performance drop on the KADID-10k dataset due to the diverse distortion types involved. However, our method still outperforms these models by a large margin, demonstrating its high effectiveness. 
Compared to the vanilla prompt used in Q-Bench, our method while utilizing the same base LMM (mPLUG-Owl3), achieves performance gains across all five datasets. 

In comparison to training-free methods, training-based models generally deliver superior results, benefiting from the quality assessment knowledge learned from large-scale datasets. However, due to the fact that KonIQ-10k only contains authentic distortions, these models often underperform on unseen distortions when tested on datasets involving unseen distortions (e.g., TReS: 0.771 on LIVE Challenge vs 0.468 on KADID-10k, MUSIQ: 0.788 on LIVE Challenge vs 0.630 on AGIQA-3k), highlighting the overfitting dilemma during training. In contrast, our approach improves the LMM in a training-free manner, providing superior generalization capability across authentic, systemic, and AI-generated distortions.\\

\noindent\textbf{Visualization.}  To verify the effectiveness of our method, we visualize our predicted results alongside the LMM predictions using the Q-Bench prompt on the KADID-10k dataset (first two rows) and the AIGC-3k dataset (second two rows).

As shown in Fig.~\ref{fig:lmmimg}, we can observe that: 1) Despite comparable distortions and closely aligned MOS distributions in the first two rows, the LMM without our debiasing enhancement consistently assigns higher quality ratings to images in the first row over the second. This observation underscores the model's high reliance on semantic content rather than low-level clues for quality assessment, revealing the presence of perceptual bias. 2) The bias varies by semantic content, affecting both natural and AI-generated images. In contrast, our debias strategy could effectively mitigate such bias, resulting in predictions that are more consistent with human ratings (\ie, MOSs).

\subsection{Generalization on other LMMs} In our method, we adopted the multimodal model mPLUG-Owl3 as our foundation model. To demonstrate the generalization capability of our enhancement strategy, we further validate it on another four LMMs inducing: BakLLaVA~\cite{liu2024improved}, Qwen-VL~\cite{bai2023qwen}, LLaVA-OneVision~\cite{li2024llavaonevisioneasyvisualtask}, LLaVA-Interleave~\cite{li2024llavanextinterleavetacklingmultiimagevideo} and  DeepSeek-VL2~\cite{wu2024deepseekvl2mixtureofexpertsvisionlanguagemodels}. As shown in Table~\ref{tab:lmm},  we could observe a consistent average performance gains can be achieved. Notably, the improvements observed on the LIVE Challenge and AGIQA-3k datasets suggest that semantic bias is widespread across diverse content types and such bias can be mitigated by our method efficiently.  The promising generalization capability highlights the transformative potential of our bias-mitigation strategy and opens up exciting new avenues for developing training-free enhancement methods to fully harness the potential of LMMs for unseen tasks.

\subsection{ Ablation Studies}
Three main components are designed in our debias scheme: 1) the conditional images, 2) the instructive prompt, and 3) the aggregation scheme. To verify the effectiveness of each component, we ablate each component from our scheme and verify their effectiveness as follows.\\

\begin{table}[htbp]
  \centering
  \caption{Ablation study of instructive prompt.}
  \resizebox{0.45\textwidth}{!}{
    \begin{tabular}{c|cc|cc|cc}
    \toprule
    \multirow{2}[4]{*}{Prompt} & \multicolumn{2}{c|}{LIVE Challenge} & \multicolumn{2}{c|}{KonIQ-10k} & \multicolumn{2}{c}{AGIQA-3k} \\
\cmidrule{2-7}          & SRCC↑ & PLCC↑ & SRCC↑ & PLCC↑ & SRCC↑ & PLCC↑ \\
    \midrule
    T1    & 0.784  & 0.762  & 0.805 & 0.816 & 0.703  & 0.682  \\
    T2    & 0.785  & 0.762  & 0.813 & 0.847 & 0.705  & 0.702  \\
    T3    & 0.741  & 0.730 &  0.811 & 0.845 & 0.672  & 0.686 \\
    \midrule
    Q-Debias & \textbf{0.794}  & \textbf{0.818}  & \textbf{0.838}  & \textbf{0.863}  & \textbf{0.717}  & \textbf{0.753}  \\
    \bottomrule
    \end{tabular}}%
 \label{tab4}

\end{table}%

\noindent\textbf{Study of Conditional Images.} In our method, distortion types are carefully selected to degrade image quality while preserving semantic content. Specifically, we explore five types of distortions for conditional image construction: \textbf{blur, noise, adverse weather conditions, brightness adjustment, and saturation modification}. For blur distortion, we consider zoom blur, motion blur, and Gaussian blur, while for noise distortion, we examine Gaussian noise and spatter noise. Additionally, we synthesize significant snow, frost, and fog distortions, as humans can still recognize objects in images captured under adverse weather conditions. 
The study results are summarized in Table~\ref{tab:dst}.

From Table~\ref{tab:dst}, we observe that all distortion types enhance AI-generated images in the AGIQA-3k dataset, whereas only a subset (\eg, spatter noise and saturation modification) improves quality prediction for natural images in the LIVE Challenge dataset. This suggests that perception bias in AI-generated images is more pronounced, making performance gains more detectable. A potential explanation is that LMMs have been exposed to significantly fewer AI-generated images compared to natural images during training due to the vast historical disparity in dataset sizes. Moreover, different distortion types exhibit varying effectiveness in bias mitigation, underscoring the importance of careful distortion design, as bias levels are highly dependent on image semantics. To develop a generalized and effective bias mitigation strategy, we further explore potential combinations of the examined distortions. Given the exponential growth in possible combinations, we evaluate a four-distortion combination scheme, selecting the most effective distortion from each category. The results demonstrate that incorporating diverse distortion types yields the highest performance gains, highlighting their complementary roles in mitigating bias.

In our approach, distortions are applied directly to the query image to generate degraded versions while maintaining semantic consistency. To assess the necessity of \textbf{semantic consistency}, we further investigate the use of conditionally distorted images with mismatched semantics. Specifically, we construct an additional conditional image set by applying the four selected distortions to open-source images that do not share semantic content with those in the LIVE Challenge and AGIQA-3k datasets. For each quality inference, we randomly select four low-quality images from this set as conditional images. The results, presented in Exp. 14 of Table~\ref{tab:dst}, show a significant performance drop when semantically inconsistent images are used, reinforcing the critical role of semantic alignment in bias mitigation. This finding highlights that bias is highly semantic-specific—leveraging conditionally degraded images that align with the query image semantics enables more accurate bias estimation and ultimately improves quality prediction.

It is important to note that our four-distortion combination represents just one potential bias mitigation strategy. Further exploration of different distortion combinations may yield superior performance. Nonetheless, our findings provide a strong foundation for refining and optimizing bias mitigation techniques in image quality assessment.
\\

\noindent\textbf{Study of Instructive Prompt.} In our method, the prompt serves an instructive role for the LMM, facilitating the propagation of bias mitigation from the conditional images to the query image. To verify its effectiveness, we compare our method against three prompt variants: \textbf{(T1)} Prompt Replacement: The entire prompt is replaced with ``{\fontfamily{cmtt}{\selectfont{\textcolor{red}{Rate the quality of the second image. Good or poor?}}}}”  \textbf{(T2)} Bias Exposure Ablation: The phrase ``The visual quality of the first image is poor” is removed from our prompt to examine the role of bias exposure, leading to the second prompt: ``{\fontfamily{cmtt}{\selectfont{\textcolor{red}{How about the visual quality of the second image? Good or poor?}}}}" \textbf{(T3)} Bias Mitigation Propagation Ablation: We delete the ``How about" from the prompt to assess the impact of bias mitigation propagation to the query image, resulting in the third prompt: ``{\fontfamily{cmtt}{\selectfont{\textcolor{red}{The visual quality of the first image is poor. Rate the visual quality of the second image. Good or poor?}}}}" As shown in Table~\ref{tab4}, the results reveal that: 1) Without our instructive prompt, even with the conditional images provided, bias cannot be effectively mitigated. 2) Without the explicit indication that the conditional images are of poor quality, the LMM fails to recognize the bias exposure, resulting in a marked performance drop. 3) The phrase ``How about” suggests that the LMM should infer the query image’s quality based on the prior understanding that the conditional images are of poor quality. Without this phrase, the propagation of bias mitigation from the conditional images to the query image weakens noticeably. The best results are achieved when the full prompt is included, demonstrating the necessity of each instruction in our prompt.\\

\begin{table}[htbp]
  \centering
  \caption{Ablation study of different aggregation schemes. }
   \resizebox{0.45\textwidth}{!}{
    \begin{tabular}{c|cc|cc|cc}
    \toprule
    \multirow{2}[4]{*}{Aggregation Scheme} & \multicolumn{2}{c|}{LIVE Challenge} & \multicolumn{2}{c|}{KonIQ-10k} & \multicolumn{2}{c}{AGIQA-3k} \\
\cmidrule{2-7}          & SRCC↑ & PLCC↑ & SRCC↑ & PLCC↑ & SRCC↑ & PLCC↑ \\
    \midrule
    Average & 0.789 & 0.754 & 0.824 & 0.838 & 0.711 & 0.691 \\
   Quality Similarity & 0.785 & 0.740  & 0.817 & 0.827 & 0.710  & 0.683 \\
    Winner-Takes- All & 0.632 & 0.491 & 0.750  & 0.660 & 0.623 & 0.529 \\
    \midrule
    Q-Debias & \textbf{0.794}  & \textbf{0.818}  & \textbf{0.838}  & \textbf{0.863}  & \textbf{0.717}  & \textbf{0.753} \\
    \bottomrule
    \end{tabular}%
    }
 \label{tab5}%
\end{table}%
\noindent\textbf{Study of Aggregation Scheme.} To aggregate the quality scores derived from different conditional images, we introduce a semantic similarity aggregation strategy. To assess its effectiveness, we compare our method with four alternative schemes: \textbf{1) Average Aggregation:} Each of the four quality scores is assigned an equal weight during aggregation. \textbf{2) Quality Similarity Aggregation:} We utilize the widely adopted FR-IQA model, LPIPS \cite{zhang2018perceptual}, to measure the quality similarity between the query image and each of its conditional images. These quality similarity scores are then treated as weights for the aggregation.
\textbf{3) Winner-Takes-All:} The final quality score is determined solely by the quality score obtained from the conditional image that exhibits the highest semantic similarity to the query image. The results are presented in Table~\ref{tab5}, which reveal that: 1) The Average scheme results in a noticeable performance drop, suggesting that uniform weighting fails to account for the bias variations across different types of distortions. 2) The Quality Similarity scheme is also ineffective. The possible reason may lie that a higher quality similarity score does not always correspond to a higher semantic recognition for the LMM, due to perceptual discrepancy between the LMM and the human visual system. 3) The Winner-Takes-All scheme, though commonly used for score aggregation, demonstrates suboptimal performance as it fails to adequately capture the nuanced contributions of different conditional images. In comparison, our semantic similarity aggregation scheme delivers the best performance across, demonstrating superior generalization  on diverse image distortions.

\section{Conclusion}
In this paper, we propose a training-free scheme to enhance the LMM in the IQA task. In particular, the perception bias that the LMM infers image quality highly relies on image semantics is mitigated by introducing conditional images in the prompt. These conditional images share the similar semantics as the query image but experience degraded quality. By instructing the LMM to align its quality ratings on those conditional images, the alignment in turn forces the LMM to rectify their judgment on the query image. Experimental results on images with different distortions verify the effectiveness of our method, and the generalization capability of our scheme across other LMMs highlights the potential for advanced prompt designs to fully leverage LMM knowledge for unseen tasks.

\section{Acknowledgments}
The work was supported by the National Natural Science Foundation of China under Grant No. 62401214.

\bibliography{aaai2026}

\end{document}